\documentclass[a4paper,conference]{IEEEtran}

\usepackage{cite}
\usepackage{bbding}
\usepackage[colorlinks,
linkcolor=black,
anchorcolor=black,
citecolor=black
]{hyperref}

\usepackage{multirow}
\usepackage{color}

\ifCLASSINFOpdf
\usepackage[pdftex]{graphicx}

\else

\fi

\usepackage{amsmath}
\usepackage{amsfonts}
\usepackage{bm}

\usepackage[marginal]{footmisc}
\IEEEoverridecommandlockouts

\DeclareRobustCommand*{\IEEEauthorrefmark}[1]{%
	\raisebox{0pt}[0pt][0pt]{\textsuperscript{\footnotesize\ensuremath{#1}}}}

\begin{document}

\title{Learning Feature Fusion for Unsupervised Domain Adaptive Person Re-identification}

\author{\IEEEauthorblockN{Jin Ding\IEEEauthorrefmark{1},
Xue Zhou\IEEEauthorrefmark{1}\IEEEauthorrefmark{2}\IEEEauthorrefmark{3}}\thanks{
\rule[0em]{4cm}{0.05em} }\thanks{\quad${}^{3}$Corresponding author, Email: zhouxue@uestc.edu.cn.}
\IEEEauthorblockA{\IEEEauthorrefmark{1}School of Automation Engineering,\\
University of Electronic Science and Technology of China (UESTC),
Chengdu, China}
\IEEEauthorblockA{\IEEEauthorrefmark{2}Shenzhen Institute for Advanced Study, UESTC, Shenzhen, China}
}

\maketitle

\begin{abstract}
Unsupervised domain adaptive (UDA) person re-identification (ReID) has gained increasing attention for its effectiveness on the target domain without manual annotations. Most fine-tuning based UDA person ReID methods focus on encoding global features for pseudo labels generation, neglecting the local feature that can provide for the fine-grained information. 
To handle this issue, we propose a Learning Feature Fusion (\bm{$LF^{2}$}) framework for adaptively learning to fuse global and local features to obtain a more comprehensive fusion feature representation. 
Specifically, we first pre-train our model within a source domain, then fine-tune the model on unlabeled target domain based on the teacher-student training strategy. 
The average weighting teacher network is designed to encode global features, while the student network updating at each iteration is responsible for fine-grained local features. By fusing these multi-view features, multi-level clustering is adopted to generate diverse pseudo labels. 
In particular, a learnable Fusion Module (\bm{$FM$}) for giving prominence to fine-grained local information within the global feature is also proposed to avoid obscure learning of multiple pseudo labels. 
Experiments show that our proposed \bm{$LF^{2}$} framework outperforms the state-of-the-art with 73.5\% mAP and 83.7\% Rank1 on Market1501 to DukeMTMC-ReID, and achieves 83.2\% mAP and 92.8\% Rank1 on DukeMTMC-ReID to Market1501. 

\end{abstract}
\IEEEpeerreviewmaketitle

\section{Introduction}

Person Re-identification (ReID) aims to identify a query that appears in one camera from a large-scale gallery set captured by other non-overlapping cameras or the same camera at different times\cite{zheng2016person}. This task has received more and more attention since it plays an essential role in widespread fields such as video surveillance, intelligent albums, urban road traffic, etc.
Although supervised person ReID methods\cite{wang2018learning,sun2018beyond,luo2019strong,zheng2019joint,zhang2020relation} have achieved satisfactory performance, they can not meet the practical needs due to time-consuming labelling when facing a new target domain. 

\begin{figure}[t]\centering
	\includegraphics[scale=0.15]{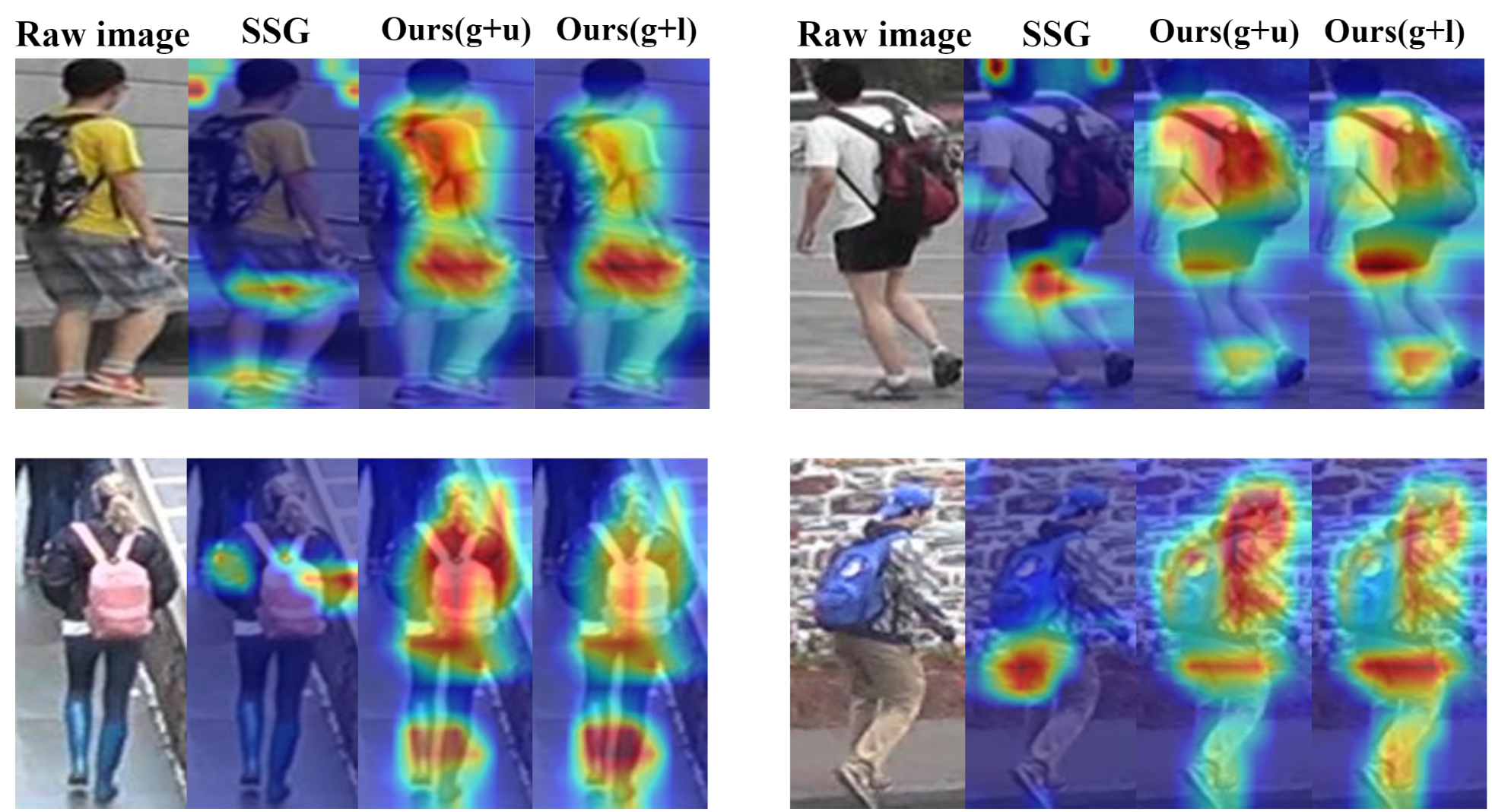}
	\caption{Visualization of the feature maps. It can be seen that the SSG method pays more attention to some irrelevant information (\textit{e.g.} backgrounds) while neglects the important characteristics (\textit{e.g.} clothes and schoolbags). However, our model greatly improves this situation. 
	``\textbf{Ours(g+u)}" and ``\textbf{Ours(g+l)}" denote the visualization results of fusing the \textbf{upper} and \textbf{lower} feature maps with the \textbf{global} feature map, respectively.}
	\label{FIG:ssg}
	\vspace{-2mm}
\end{figure}

A feasible solution is to adapt the model trained on a source domain with labels to an unlabeled target domain, known as Unsupervised domain adaptive (UDA) person ReID. Yet it is still a challenging problem due to existing data distribution gap and non-overlapping identities between source and target domains. 
Most existing UDA person ReID methods\cite{fu2019self,zhai2020ad,ji2020attention,ge2020mutual,zhai2020multiple,zheng2021group,zheng2021online,zheng2021exploiting,dai2021dual}, termed as ``fine-tuning'', firstly pre-train the model on the labeled source domain and then adopt clustering algorithms or similarity measurements for pseudo labels generation on the unlabeled target domain to fine-tune the model. 
To this end, some methods\cite{zhang2019self,song2020unsupervised,NEURIPS2020_821fa74b,yang2021progressive} use a single feature extractor to produce the global target-domain features for further pseudo labels generation, leading to many unreliable pseudo labels. Other methods\cite{ge2020mutual, zhai2020multiple,liu2021graph} adopting the teacher-student framework iteratively learn an average weighting model to generate the global feature for obtaining more reliable pseudo labels.
Despite the significant success, these fine-tuning methods focus on extracting global features that only contain coarse semantic information and neglect the local feature being capable of providing the fine-grained information. 

Self-similarity Grouping (SSG)\cite{fu2019self} is the first to consider both global and local features in UDA person ReID. However, we found that SSG has two issues. Firstly, using a single network to extract features for clustering is susceptible to generate many noisy pseudo labels. Secondly, SSG implements clustering based on global and local features independently. This may lead to an unlabeled sample with multiple completely different pseudo labels, so the model will not be able to clearly classify it to which identity during the training, which is defined as \textit{obscure learning} in this paper.
As shown in Fig.\ref{FIG:ssg}, SSG is prone to focus on some irrelevant information, resulting in the failure to comprehensive feature learning.

\begin{figure*}[t]\centering
	\includegraphics[width=0.95\textwidth]{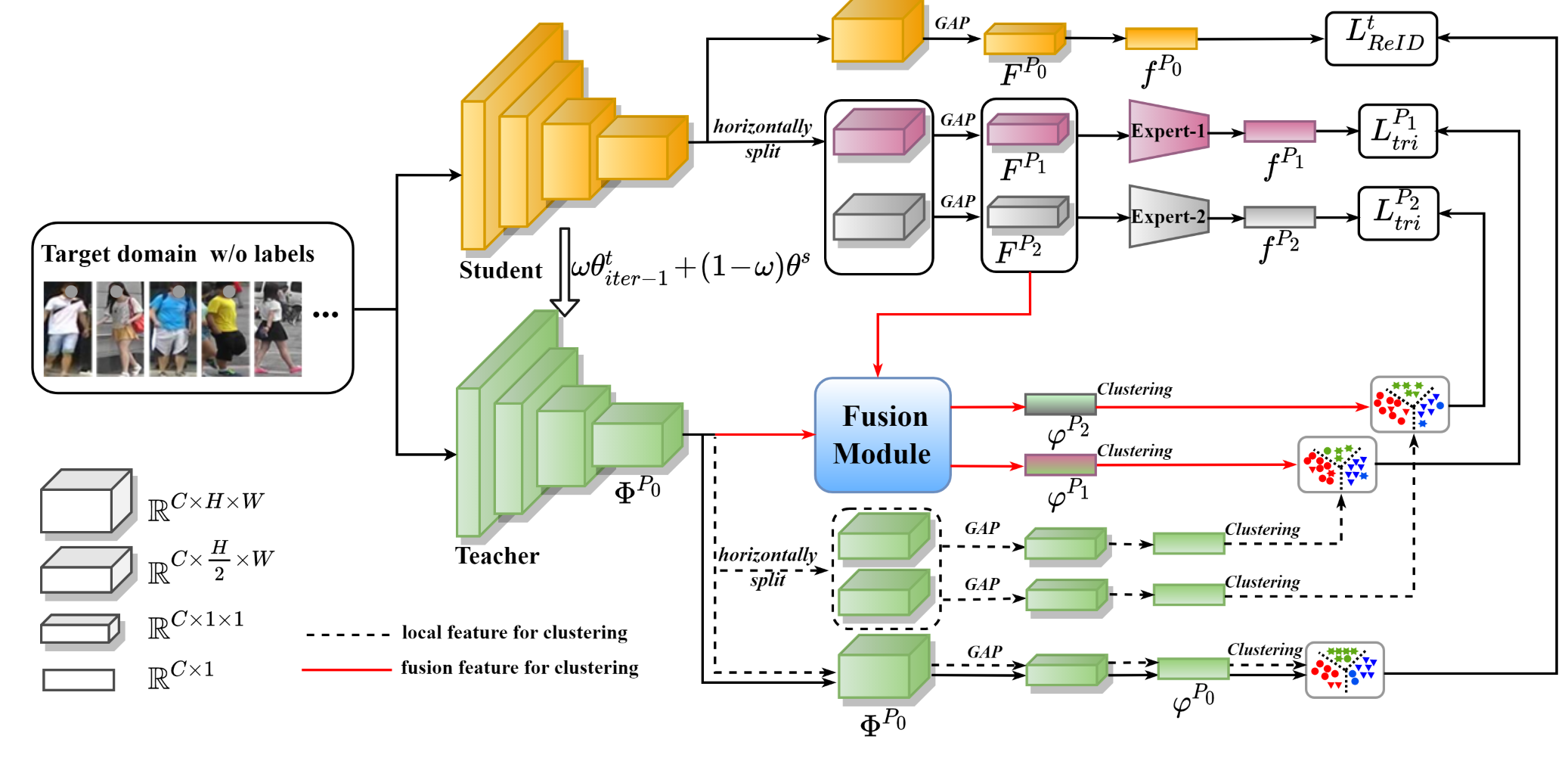}
	\caption{Overall of our Learning Feature Fusion ($LF^{2}$) framework. GAP denotes global average pooling. During the target-domain training, there are two steps: 1) Learning feature fusion. Both global feature map $\Phi^{P_{0}}$ and local feature maps $F^{P_{j}}, j\in\{1,2\}$ are forwarded to the Fusion Module ($FM$), as shown by the red lines. We evaluate the effect of $FM$ by replacing the red lines with the dotted ones. 2) Model optimization. The student network is optimized under the supervision of pseudo labels while the teacher network is updated by the temporal average of the student ones. Only teacher network is adopted for inference.}
	\label{FIG:2}

\end{figure*}

In this paper we propose a Learning Feature Fusion ($LF^{2}$) framework to cope with the above mentioned problems. 1) We design a mean-teacher\cite{NIPS2017_68053af2} based framework to iteratively learn multi-view features. This framework can build multiple clusters in order to refine the noisy pseudo labels. 2) With the help of a learnable global-to-local Fusion Module (\textbf{$FM$}), our method mitigates the obscure learning. 
Specifically, we firstly pre-train our model within a source domain. Then, inspired by the teacher-student framework\cite{NIPS2017_68053af2,zhang2018deep}, we design the overall structure of our $LF^{2}$ framework, as shown in Fig.\ref{FIG:2}.
To obtain diverse pseudo labels, we fuse the local features encoded from the student network and the global feature encoded from the teacher network, and build multiple clusters as shown by the red lines in Fig.\ref{FIG:2}.
With a learnable $FM$, the local features will be adaptively fused with the global feature. As a result, not only can the fusion features learn more comprehensive representations, but also induce more consistent pseudo labels to avoid obscure learning. 

The contributions of this paper are summarized as follows.
\begin{itemize}
	\item[$\bullet$] We build a Learning Feature Fusion ($LF^{2}$) framework upon a pair of teacher-student network. Wherein multi-view features are adaptively fused for multi-level clustering, which aims to obtain diverse pseudo labels.
	\item[$\bullet$]To learn more comprehensive representations and to avoid obscure learning of multiple pseudo labels, we design a learnable Fusion Module ($FM$) which focuses on the fine-grained local information in global feature.
	\item[$\bullet$] Experiments conducted on two common UDA ReID settings show that our method achieves significant performance gain over the state-of-the-arts.
\end{itemize}

\section{Related work}
To further exploit the existing labeled person ReID datasets (domains) to adapt to unlabeled target domains, many UDA person ReID methods have been proposed in recent years.
These methods can be categorized into three types: Generative Adversarial Network (GAN) transferring based methods\cite{deng2018image,li2019cross}, joint learning based methods\cite{zhong2019invariance,li2020joint, wang2020unsupervised,dai2021idm} and fine-tuning based methods\cite{fu2019self,NEURIPS2020_821fa74b,yang2021progressive,ge2020mutual,zhai2020multiple,zhai2020ad,ji2020attention,zheng2021group,zheng2021online,zheng2021exploiting,dai2021dual}. 

\textbf{GAN transferring} based methods apply GANs to generate images with target-domain styles while preserving the identities of source domain as far as possible. Deng \textit{et al.}\cite{deng2018image} propose a Similarity Preserving Generative Adversarial Network (SPGAN) to preserve the self-similarity and domain-dissimilarity during image-image translation. \textbf{Joint learning} based methods focus on exploring the underlying representation between source and target domain. Zhong \textit{et al.}\cite{zhong2019invariance} propose to use labeled source data and unlabeled target data to obtain up-to-date representation with two components respectively.
\textbf{Fine-tuning} based methods usually consist of two stages: pre-training the model on the labeled source domain and optimizing it on the unlabeled target domain with pseudo labels generated by clustering algorithm or similarity measurement. To this end, Song \textit{et al.}\cite{song2020unsupervised} propose a self-training scheme to iteratively extract features for clustering and minimize the loss functions with a single network. The approach proposed by Zhai \textit{et al.}\cite{zhai2020multiple} focuses on multiple pairs of teacher-student networks and uses the global feature of the teacher networks to refine the noisy pseudo labels.  However, these fine-tuning methods omit the potential pseudo labels refinery from global to local features existing in the target-domain training samples. Fu \textit{et al.}\cite{fu2019self} adopt local and global features for independently building multiple clusters, which brings about the obscure learning.

Compared with them, our $LF^{2}$ framework uses target-domain features from global to local for learning fusion and building multiple clusters. 
It ensures the fusion features without obscurity and learns more comprehensive representations.

\section{Proposed approach}
\subsection{Overview}
\textbf{Problem definition.} We denote the labeled source domain data as $\mathcal{S}=\{(x_{i}^{s},y_{i}^{s})\}_{i=1}^{N_{s}}$, where $x_{i}^{s}$ and $y_{i}^{s}$ denote the \textit{i}-$th$ source image and its unique identity label, $N_{s}$ denotes the number of source-domain sample images. We denote the unlabeled target domain data as $\mathcal{T}=\{x_{i}^{t}\}_{i=1}^{N_{t}}$, where $x_{i}^{t}$ denotes the \textit{i}-$th$ target sample. Identity labels are not available for the images in the target domain dataset.
It is worth noting that the identities of these domains are non-overlapping. 
The goal of UDA person ReID is to transfer the knowledge from the source domain $\mathcal{S}$ to the target domain $\mathcal{T}$. We propose a $LF^{2}$ framework in a fine-tuning manner to achieve this goal. 

\textbf{Methodology.} We adopt the two-stage fine-tuning method: pre-training the model on the source domain and fine-tuning it on the target domain. A pair of teacher-student network is adopted in our method. 
The student network is first pre-trained on the source domain in a supervised manner and the pre-trained parameters will be copied to the teacher network. Two steps will be iteratively adopted during target-domain fine-tuning, as shown in Fig.\ref{FIG:2}. 
1) Learning feature fusion. Both student network's local feature maps and teacher network's global feature map are forwarded to the $FM$ for adaptively learning fusion. The global feature $\varphi^{P_{0}}$ and the fusion features $\varphi^{P_{j}},j\in\{1,2\}$ are used for clustering, thereby establishing multiple new datasets with pseudo labels. 2) Model optimization. The student network is optimized via loss functions while the teacher network is updated by the temporal average of the student ones.
Details are referred in the following parts. 

\subsection{Source-domain pre-training}
We first train our model on the source domain data $\mathcal{S}$ in a supervised manner. Similar to most supervised ReID methods, this phase mainly contains two loss functions: the classification loss $L_{cls}$ and the triplet loss $L_{tri}$\cite{hermans2017defense}. With its corresponding class label $y_{i}^{s}$ of the \textit{i}-$th$ source-domain sample $x_{i}^{s}$, the classification and triplet losses can be defined as follows:
\begin{equation}
	{L}_{cls}^{s} = \frac{1}{N_{s}} \sum_{i=1}^{N_{s}}L_{ce}(C^{s}(f(x_{i}^{s})), y_{i}^{s}),
	\label{L_cls_source}
\end{equation}
where $f(x_{i}^{s})$ is the feature of source image $x_{i}^{s}$ and $L_{ce}$ is cross entropy loss, $C^{s}$ is a fully connected layer for classification: $f(x_{i}^{s})\rightarrow\{1,2,...,M_{s}\}$, $M_{s}$ is the number of identities of source domain.
\vspace{-1mm}
\begin{equation}
	\begin{split}
		{L}_{tri}^{s} \!=\! -\frac{1}{N_{s}} \sum_{i=1}^{N_{s}}\max(0, m \!+\!\left\|f(x_{i}^{s})\!-\! f(x_{i,+}^{s})\right\|_2\!-\! \\ \left\|f(x_{i}^{s})\!-\! f(x_{i,-}^{s})\right\|_2),
		\label{L_tri_source}
	\end{split}
\end{equation}
where $x_{i}^{s}$, $x_{i,+}^{s}$, $x_{i,-}^{s}$ denote the anchor, positive and negative samples, respectively. $m$ is the margin and $\left\|\cdot\right\|_2$ is the $L_{2}$ distance. The overall loss is therefore calculated as:
\begin{equation}
	{L}_{ReID}^{s}=L_{cls}^{s}+\lambda{L}_{tri}^{s}.
\end{equation}

With the source-domain ground-truth labels and a balance hyper-parameter $\lambda$, the model shows good performance. However, this performance will dramatically drop if the model is directly applied on the unlabeled target domain.

\subsection{Target-domain fine-tuning}
In this phase, we use the pre-trained model for fine-tuning. We present our framework by introducing two steps: learning feature fusion and model optimization.

\textbf{Learning feature fusion.} 
Our motivation is to make the model adaptively fuse the global and local features, resulting in features focused on both global and local information. In addition, building multiple clusters with global and fused features induces more consistent pseudo labels to avoid obscure learning.

To refine the noisy pseudo labels, we choose a pair of teacher-student network based on mean-teacher\cite{NIPS2017_68053af2} as our baseline. We feed the same unlabeled image of the target domain to the teacher and student networks. The student network's parameters $\theta^{s}$ will be updated via back-propagation during the target-domain training while the teacher network's parameters $\theta^{t}$ are computed as the temporal average of the student parameters $\theta^{s}$, which can be calculated as: 
\begin{equation}
	\label{temporal}
	\theta_{iter}^{t}=\omega \theta_{iter-1}^{t}+(1-\omega)\theta^{s},
\end{equation}
where $iter$ is the current iteration and $\omega$ is the temporal ensemble momentum with the range $\!\left[0,1\right)$. 

To obtain the fusion features, we horizontally partition the last global feature map of the student network into \textit{K} parts and obtain the results $F^{P_{j}}, j\in \{1,2,3,...,K\}$ after global average pooling, while the last global feature map $\Phi^{P_{0}}$ of the teacher network will not be split. In our framework, \textit{K} is empirically set to 2, as shown in Fig.\ref{FIG:2}. 
$F^{P_{1}}, F^{P_{2}}$ of the student network and $\Phi^{P_{0}}$ of the teacher network are selected for adaptively learning feature fusion by the $FM$ module with learnable parameters. There are following two reasons. 
1) The teacher network gathers the learned knowledge from the student network in each iteration meanwhile maintains previous knowledge to update its parameters as shown by Eq.(\ref{temporal}). Thus, the global feature map of the teacher network can focus on the overall information of the image during the whole training, which is more comprehensive to represent the global characteristics of the training sample.
2) With the optimization of loss functions, the student network pays attention to diverse characteristics in each iteration, which means that the local feature maps used for learning fusion can avoid neglecting some important local fine-grained information.

\begin{figure}[t]\centering
	\includegraphics[width=0.45\textwidth]{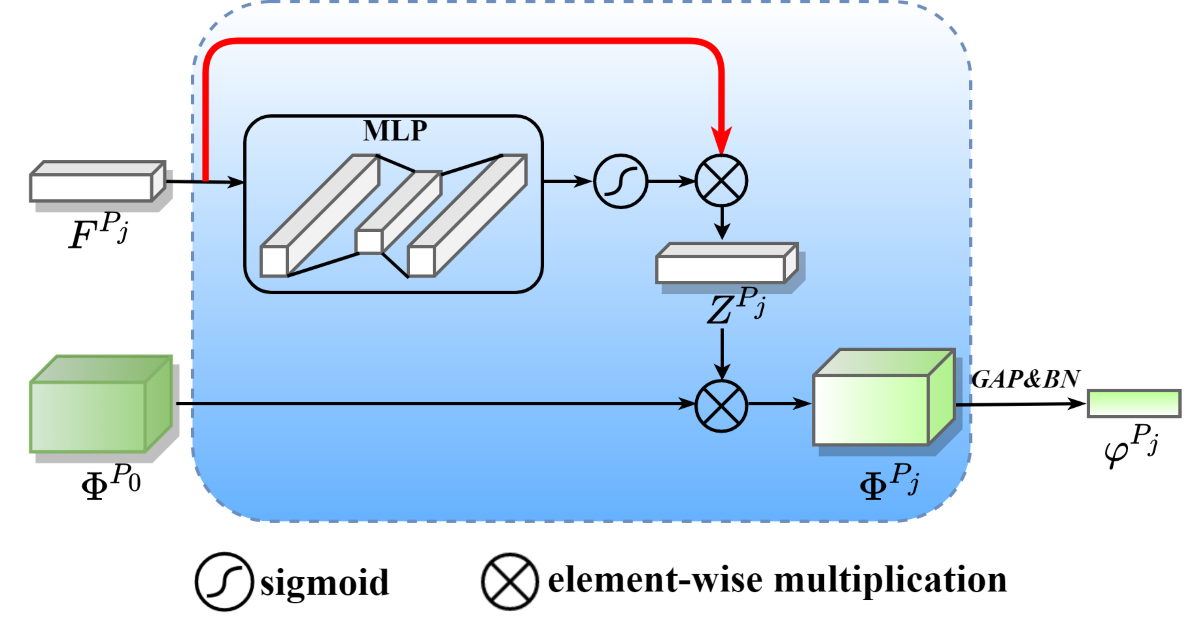}
	\caption{The structure of Fusion Module($FM$). As illustrated, the local feature map $F^{P_{j}},j\in\{1,2\}$ is first forwarded to a MLP for adaptively learning fusion. The red line is a residual structure to obtain the learned attention map $Z^{P_{j}}$. Then, the fusion feature map $\Phi^{P_{j}}$ is obtained by the element-wise multiplication with $Z^{P_{j}}$ and the global feature map $\Phi^{P_{0}}$.}
	\label{FIG:3}
	\vspace{-4mm}
\end{figure}

Inspired by CBAM\cite{woo2018cbam}, we design the $FM$, as shown in Fig.\ref{FIG:3}. We first adopt $F^{P_{j}}$ with spatial information of the student network's local feature to a multi-layer perceptron (MLP) with one hidden layer for learning fusion. The size of the hidden activation is set to $\mathbb{R}^{\frac{C}{r}\times1\times1}$, where $r$ is the reduction ratio and $C$ represents the channel number. To highlight the learned fusion knowledge of the local feature maps, we design a residual structure to aggregate $F^{P_{j}}$ and the output of MLP to obtain the attention map $Z^{P_{j}} \in \mathbb{R}^{C\times 1 \times 1}$, as shown by the red line in Fig.\ref{FIG:3}. Then, we merge the global feature map $\Phi^{P_{0}} \in \mathbb{R}^{C\times H \times W}$ of the teacher network and $Z^{P_{j}}$ by using element-wise multiplication with spatial dimension broadcast, where $H$ and $W$ represent the height and width. 
The overall process can be summarized as: 
\begin{equation}
	\begin{aligned}
		\Phi^{P_{j}}
		& = \sigma (Z^{P_{j}}) \otimes \Phi^{P_{0}} \\
		& = \sigma (MLP(F^{P_{j}}) \otimes F^{P_{j}} ) \otimes \Phi^{P_{0}} \\
		& = \sigma (W_{2}(ReLU(W_{1}(F^{P_{j}})))\otimes F^{P_{j}}) \otimes \Phi^{P_{0}},
	\end{aligned}
\end{equation}
where $j \in \{1,2\}$, $\sigma$ denotes the sigmoid function, $W_{1} \in \mathbb{R}^{ \frac{C}{r} \times C }$, $W_{2} \in \mathbb{R}^{C \times \frac{C}{r}}$. 
Note, each local feature map $F^{P_{j}}$ has its own MLP for learning feature fusion. 

Both fusion features $\varphi^{P_{j}}, j \in \{1,2\}$ and global feature $\varphi^{P_{0}}$ are finally utilized to predict the pseudo labels by cluster algorithm, \textit{e.g.}, k-means or DBSCAN\cite{ester1996density}. As a result, each target-domain image $x_{i}^{t}$ has $K+1$ pseudo labels. The target domain data can be denoted as: $\mathcal{T}=\{(x_{i}^{t}, \hat{y}_{i,j}^{t}|_{j=0}^{K})\}_{i=1}^{N_{t}}$, where $\hat{y}_{i,j}^{t}|_{j=0}^{K} \in \{1,2,...,M_{t,j}\}$ denotes that the pseudo label $\hat{y}_{i,j}^{t}$ of the image $x_{i}^{t}$ is from the cluster result $\hat{Y}_{j}=\{\hat{y}_{i,j}^{t}| i=1,2,...,N_{t}\}$ of the feature $\varphi^{P_{j}}$, $M_{t,j}$ denotes the number of identities in the $\hat{Y}_{j}$. 

\textbf{Model optimization.} After obtaining multiple pseudo labels, we can get $K+1$ new target-domain datasets for training the student network. To ensure the fusion features without spatial deviation in the learning process, we exploit the counterpart features of the student network for optimizing model. 
In particular, for each local feature map $F^{P_{j}}$ of the student network, we design \textbf{Expert-j} that is capable of aligning the student network's local feature and the fusion feature(\textit{i.e.}, $f^{P_{j}}$ and $\varphi^{P_{j}}$).
For the sake of simplicity, we omit the superscripts of $x_{i}^{t}$ and $\hat{y}_{i,j}^{t}$ in the following text without confusion. 

For the local features $f^{P_{j}}$, we use the pseudo labels generated by corresponding fusion feature $\varphi^{P_{j}}$ to calculate the softmax triplet loss. 
The softmax triplet loss is defined as: 
\begin{equation}
	{L}_{tri}^{P_{j}}=-\frac{1}{N_{t}}\sum_{i=1}^{N_{t}}\log\mathcal{H}_{j}(x_{i}|\theta^{s}), j\in\{1,2\},
	\label{tri_g}
\end{equation}
where $\mathcal{H}_{j}(x_{i}|\theta^{s})$ is formulated as: 
\begin{equation}
	\frac{e^{\left\|f^{P_{j}}(x_{i}|\theta^{s})\!-\! f^{P_{j}}(x_{i,-}|\theta^{s})\right\|_2}}{e^{\left\|f^{P_{j}}(x_{i}|\theta^{s})\!-\! f^{P_{j}}(x_{i,+}|\theta^{s})\right\|_2}\!+\! e^{\left\|f^{P_{j}}(x_{i}|\theta^{s})\!-\! f^{P_{j}}(x_{i,-}|\theta^{s})\right\|_2}}.
	\label{h_formulated}
\end{equation}
$x_{i,+}$ and $x_{i,-}$ are the hardest positive and negative samples of the anchor target-domain image $x_{i}$, respectively. 

\begin{table*}[t]\centering
	\renewcommand\arraystretch{1.2}
	\caption{Comparison of the proposed \textbf{$LF^{2}$} method with state-of-the-art methods. ``\textbf{--}'' indicates the value is not provided in the corresponding paper. \textbf{Top three} performance values are highlighted in \textcolor{red}{\textbf{red}}, \textcolor{blue}{\textbf{blue}} and \textcolor[RGB]{255,140,0}{\textbf{orange}} colors respectively.}
	\label{table_1}
	\begin{tabular}{ll|ccccccccc}
		\hline
		\multicolumn{1}{c|}{\multirow{2}{*}{Categories}} & \multicolumn{1}{|c|}{\multirow{2}{*}{Methods}} & \multicolumn{1}{|c|}{\multirow{2}{*}{Reference}} & \multicolumn{4}{|c|}{D-to-M}                                                                  & \multicolumn{4}{|c}{M-to-D}                                                                  \\
		\multicolumn{1}{c|}{}                            & \multicolumn{1}{|c|}{}                         & \multicolumn{1}{|c|}{}                           & \multicolumn{1}{c}{mAP} & \multicolumn{1}{c}{Rank1} & \multicolumn{1}{c}{Rank5} & \multicolumn{1}{c|}{Rank10} & \multicolumn{1}{c}{mAP} & \multicolumn{1}{c}{Rank1} & \multicolumn{1}{c}{Rank5} & \multicolumn{1}{c}{Rank10} \\
		\hline
		\multicolumn{1}{l|}{\multirow{2}{*}{GAN transferring}}               & SPGAN+LMP\cite{deng2018image}                                    & \multicolumn{1}{l|}{CVPR'18 }                                      & 26.7                    & 57.7                      & 75.8                      & \multicolumn{1}{c|}{82.4}                       & 26.2                    & 46.4                      & 62.3                      & 68.0                       \\ \multicolumn{1}{l|}{}
		& PDA-Net\cite{li2019cross}                                      & \multicolumn{1}{l|}{ICCV'19}                                       & 47.6                    & 75.2                      & 86.3                      & \multicolumn{1}{c|}{90.2}                       & 45.1                    & 63.2                      & 77.0                      & 82.5                       \\
		\hline
		\multicolumn{1}{l|}{\multirow{4}{*}{Joint learning}}                 & ECN\cite{zhong2019invariance}                                          & \multicolumn{1}{l|}{CVPR'19}                                       & 43.0                    & 75.1                      & 87.6                      & \multicolumn{1}{c|}{91.6}                       & 40.4                    & 63.3                      & 75.8                      & 80.4                       \\ \multicolumn{1}{l|}{}
		& MMCL\cite{wang2020unsupervised}                                         & \multicolumn{1}{l|}{CVPR'20}                                       & 60.4                    & 84.4                      & 92.8                      & \multicolumn{1}{c|}{95.0}                       & 51.4                    & 72.4                      & 82.9                      & 85.0                       \\ \multicolumn{1}{l|}{}
		& JVTC+\cite{li2020joint}                                        & \multicolumn{1}{l|}{ECCV'20}                                       & 67.2                    & 86.8                      & 95.2                      & \multicolumn{1}{c|}{97.1}                       & 66.5                    & 80.4                      & 89.9                      & 92.2                       \\  \multicolumn{1}{l|}{}
		& IDM\cite{dai2021idm}                                        & \multicolumn{1}{l|}{ICCV'21}                                       & \textcolor{blue}{\textbf{82.8}}                    & \textcolor{red}{\textbf{93.2}}                      & \textcolor[RGB]{255,140,0}{\textbf{97.5}}                      & \multicolumn{1}{c|}{\textcolor[RGB]{255,140,0}{\textbf{98.1}}}                       & \textcolor[RGB]{255,140,0}{\textbf{70.5}}                    & \textcolor{blue}{\textbf{83.6}}                      & \textcolor{blue}{\textbf{91.5}}                      & \textcolor{blue}{\textbf{93.7}}                       \\
		\hline
		\multicolumn{1}{l|}{\multirow{10}{*}{Fine-tuning}}                   & SSG\cite{fu2019self}                                          & \multicolumn{1}{l|}{ICCV'19}                                       & 58.3                    & 80.0                      & 90.0                      & \multicolumn{1}{c|}{92.4}                       & 53.4                    & 73.0                      & 80.6                      & 83.2                       \\ \multicolumn{1}{l|}{}
		& ADTC\cite{ji2020attention}                                         & \multicolumn{1}{l|}{ECCV'20}                                       & 59.7                    & 79.3                      & 90.8                      & \multicolumn{1}{c|}{94.1}                       & 52.5                    & 71.9                      & 84.1                      & 87.5                       \\ \multicolumn{1}{l|}{}
		& AD-Cluster\cite{zhai2020ad}                                   & \multicolumn{1}{l|}{CVPR'20}                                       & 68.3                    & 86.7                      & 94.4                      & \multicolumn{1}{c|}{96.5}                       & 54.1                    & 72.6                      & 82.5                      & 85.5                       \\ \multicolumn{1}{l|}{}
		& MMT\cite{ge2020mutual}                                          & \multicolumn{1}{l|}{ICLR'20}                                       & 71.2                    & 87.7                      & 94.9                      & \multicolumn{1}{c|}{96.9}                       & 65.1                    & 78.0                      & 88.8                      & 92.5                       \\ \multicolumn{1}{l|}{}
		& MEB-Net\cite{zhai2020multiple}                                      & \multicolumn{1}{l|}{ECCV'20}                                       & 76.0                    & 89.9                      & 96.0                      & \multicolumn{1}{c|}{97.5}                       & 66.1                    & 79.6                      & 88.3                      & 92.2                       \\ \multicolumn{1}{l|}{}
		& Dual-Refinement\cite{dai2021dual}                              & \multicolumn{1}{l|}{TIP'21 }                                       & 78.0                    & 90.9                      & 96.4                      & \multicolumn{1}{c|}{97.7}                       & 67.7                    & 82.1                      & 90.1                      & 92.5                       \\ \multicolumn{1}{l|}{}
		& UNRN\cite{zheng2021exploiting}                                         & \multicolumn{1}{l|}{AAAI'21}                                       & 78.1                    & 91.9                      & 96.1                      & \multicolumn{1}{c|}{97.8}                       & 69.1                    & 82.0                      & 90.7                      & \textcolor[RGB]{255,140,0}{\textbf{93.5}}                       \\ \multicolumn{1}{l|}{}
		& GLT\cite{zheng2021group}                                          & \multicolumn{1}{l|}{CVPR'21}                                       & 79.5                    & 92.2                      & 96.5                      & \multicolumn{1}{c|}{97.8}                       & 69.2                    & 82.0                      & 90.2                      & 92.8                       \\  
		\multicolumn{1}{l|}{}
		& HCD\cite{zheng2021online}                                         & \multicolumn{1}{l|}{ICCV'21}                                       & 80.0                    & 91.5                & --                        & \multicolumn{1}{c|}{--}                         & 70.1                    & 82.2                      & --                        & --                         \\  
		\multicolumn{1}{l|}{}
		& $P^{2}LR$\cite{han2021delving}                                         & \multicolumn{1}{l|}{AAAI'22}                                       & 81.0                    & 92.6                      & 97.4                        & \multicolumn{1}{c|}{\textcolor{blue}{\textbf{98.3}}}                         & \textcolor{blue}{\textbf{70.8}}                    & \textcolor[RGB]{255,140,0}{\textbf{82.6}}                      & \textcolor[RGB]{255,140,0}{\textbf{90.8}}                        & \textcolor{blue}{\textbf{93.7}}                         \\ 
		\multicolumn{1}{l|}{}
		& RDSBN+MDIF\cite{bai2021unsupervised}                                         & \multicolumn{1}{l|}{CVPR'21}                                       & \textcolor[RGB]{255,140,0}{\textbf{81.5}}                    & \textcolor{blue}{\textbf{92.9}}                      & \textcolor{blue}{\textbf{97.6}}                        & \multicolumn{1}{c|}{\textcolor{red}{\textbf{98.4}}}                         & 66.6                    & 80.3                      & 89.1                        & 92.6                         \\ 
		\cline{2-11} 
		\multicolumn{1}{l|}{}
		
		& \textbf{\bm{$LF^{2}$}(Ours)}                                         & \multicolumn{1}{l|}{This paper}                                    & \textcolor{red}{\textbf{83.2}}                    & \textcolor[RGB]{255,140,0}{\textbf{92.8}}                      & \textcolor{red}{\textbf{97.8}}                      & \multicolumn{1}{c|}{\textcolor{red}{\textbf{98.4}}}                       &\textcolor{red}{\textbf{73.5}}                     &\textcolor{red}{\textbf{83.7}}                       &\textcolor{red}{\textbf{91.9}}                       &\textcolor{red}{\textbf{94.3}}                        \\ \hline 
	\end{tabular}
\vspace{-4mm}
\end{table*}

For the global feature $f^{P_{0}}$, the cluster result $\hat{Y}_{0}$ of the global clustering feature $\varphi^{P_{0}}$ are considered as pseudo labels. Both classification loss and triplet loss are utilized for supervised learning. We define $L_{cls}^{t}$ and $L_{tri}^{t}$ as: 
\begin{equation}
	{L}_{cls}^{t}=\frac{1}{N_{t}}\sum_{i=1}^{N_{t}}L_{ce}(C^{t}(f^{P_{0}}(x_{i})), \hat{y}_{i,0}),
\end{equation}
\vspace{-2mm}
\begin{equation}
	\begin{split}
		{L}_{tri}^{t} 
		= -\frac{1}{N_{t}} \sum_{i=1}^{N_{t}}\max(0, m+\left\|f^{P_{0}}(x_{i})\!-\! f^{P_{0}}(x_{i,+})\right\|_2\!-\! \\ \left\|f^{P_{0}}(x_{i})\!-\! f^{P_{0}}(x_{i,-})\right\|_2),
	\end{split}
\end{equation}
where $C^{t}$ is the fully connected layer of the student network for classification: $f^{P_{0}}(x_{i})\rightarrow\{1,2,...,M_{t,0}\}$. 

The total loss is defined as: 
\vspace{-1mm}
\begin{equation}
	\begin{aligned}
		{L}_{total}
		&\!=\!\alpha L_{ReID}^{t} + \gamma\sum_{j=1}^{K}L_{tri}^{P_{j}}\\
		&\!=\!\alpha (L_{cls}^{t} + \lambda L_{tri}^{t}) +  \gamma\!\sum_{j=1}^{K}L_{tri}^{P_{j}},
	\end{aligned}
\end{equation}
where $\alpha$, $\lambda$ and $\gamma$ are weighting parameters. 

During the training process on the target domain, learning feature fusion and model optimization is done iteratively. The $FM$ module will be discarded and only the teacher network is used during testing. 
Specifically, the global feature map $\Phi^{P_{0}}$ of the teacher network will be partitioned into K parts. After global average pooling, K local features and the global feature are concatenated for inference. Performance of only using the global feature for inference is also experimented and the result is disappointing which is not illustrated due to the page limit.  

\begin{table}[h]
	\renewcommand\arraystretch{1.1}
	\caption{Component analysis and different number of pseudo identities ablation studies of the performance of $LF^{2}$.}
	\label{table_2}
	\centering
	\begin{tabular}{lcccc}
		\hline
		\multicolumn{1}{c}{\multirow{2}{*}{Methods}} & \multicolumn{2}{|c|}{D-to-M} & \multicolumn{2}{c}{M-to-D} \\
		\cline{2-5}
		\multicolumn{1}{c|}{} 
		& mAP         & \multicolumn{1}{c|}{Rank1}        & mAP         & Rank1        \\
		\hline
		\multicolumn{1}{l|}{Direct transfer}        & 27.8        & \multicolumn{1}{c|}{55.6}         & 26.9        & 42.6         \\
		\multicolumn{1}{l|}{Baseline(only $L_{ReID}^{t}$)}                 & 69.0        & \multicolumn{1}{c|}{86.6}         & 61.3        & 75.6         \\
		\multicolumn{1}{l|}{$LF^{2}$ w/o $FM$}              & 78.5        & \multicolumn{1}{c|}{90.5}         & 68.5        & 81.5         \\
		\hline
		\multicolumn{1}{l|}{$LF^{2}$($M_{t,j}$=500)}               & 79.9        & \multicolumn{1}{c|}{91.8}         & 68.7        & 81.7         \\
		\multicolumn{1}{l|}{$LF^{2}$($M_{t,j}$=700)}               &\textbf{83.2}             &\multicolumn{1}{c|}{\textbf{92.8}}              & 72.2        & 82.9         \\
		\multicolumn{1}{l|}{$LF^{2}$($M_{t,j}$=900)}               & 82.3        & \multicolumn{1}{c|}{92.4}         & \textbf{73.5}        & \textbf{83.7} \\
		\hline           
	\end{tabular}
	\vspace{-5.5mm}
\end{table}

\section{Experiments}
\subsection{Experimental Configurations}
\textbf{Datasets and Evaluation Metrics.} We evaluate our proposed method in two widely-used UDA person ReID settings, $i.e.$, DukeMTMC-ReID to Market1501 and Market1501 to DukeMTMC-ReID (termed as ``\textbf{D-to-M}'' and ``\textbf{M-to-D}'' in the following text). Market1501\cite{zheng2015scalable} is composed of 32,668 images of 1,501 identities from 6 camera views, for which 12,936 images of 751 identities are used for training, and the rest of images are used in the test set. DukeMTMC-ReID\cite{ristani2016performance} contains 16,522 images of 702 persons for training, 2,228 query images and 17,661 gallery images of the remaining 702 persons for testing. Cumulative Matching Characteristics (CMC) and mean Average Precision (mAP) are adopted as evaluation protocols. 

\textbf{Implementation Details.} ResNet50\cite{he2016deep} is adopted as our backbone. Referring to \cite{luo2019bag}, the last residual layer's stride size is set as 1. We resize each input image to fixed $256 \times 128$ and use some common data augmentations, $i.e.$, random horizontal flipping, random cropping, and zero padding. Random erasing\cite{zhong2020random} is only adopted during the target-domain fine-tuning. The mini-batch size is 64, including 16 randomly selected identities with 4 images of each identity. For the source-domain pre-training, the initial learning rate is 0.00035 and is decreased to its 0.1x and 0.01x at the 40th and 70th epoch in the total 80 epochs. For the target-domain fine-tuning, we totally train 80 epochs and each epoch contains 400 iterations. The learning rate is fixed to 0.00035 and we use the Adam with weight decay 0.0005. 
Inspired by \cite{ge2020mutual,zhai2020multiple} and \cite{yang2021progressive}, we utilize K-means algorithm and set the number $M_{t,j}$ of pseudo identities as 500, 700 and 900, respectively. 
The temporal ensemble momentum $\omega$ in Eq.(\ref{temporal}) is set to 0.999. The structure of the experts in the student network is implemented with FC$\rightarrow$BN, where FC is 2048d$\rightarrow$2048d. The loss weights $\alpha, \lambda, \gamma$ are set as 1,0.5,0.5 respectively. In the $FM$ module, the reduction ratio $r$ is set to 4. 
The whole model is optimized with two GeForce RTX 2080Ti GPUs.

\subsection{Comparison with the State-of-the-Art}
We compare $LF^{2}$ framework with state-of-the-art methods including: GAN transferring based methods (SPGAN+LMP\cite{deng2018image}, PDA-Net\cite{li2019cross}), joint learning based methods (ECN\cite{zhong2019invariance}, MMCL\cite{wang2020unsupervised}, JVTC+\cite{li2020joint}, IDM\cite{dai2021idm}) and fine-tuning based methods (SSG\cite{fu2019self}, ADTC\cite{ji2020attention}, AD-Cluster\cite{zhai2020ad}, MMT\cite{ge2020mutual}, MEB-Net\cite{zhai2020multiple}, Dual-Refinement\cite{dai2021dual}, UNRN\cite{zheng2021exploiting}, GLT\cite{zheng2021group}, HCD\cite{zheng2021online}, $P^{2}LR$\cite{han2021delving},  RDSBN+MDIF\cite{bai2021unsupervised}). The comparison results are shown in Table \ref{table_1}.

On \textbf{M-to-D}, $LF^{2}$ achieves the best performance compared with others. For example, $LF^{2}$ achieves 73.5\% mAP and 83.7\% Rank1. This outperforms $P^{2}LR$\cite{han2021delving} by 2.7\% in mAP accuracy and achieves the state-of-the-art performance. Both MMT\cite{ge2020mutual} and MEB-Net\cite{zhai2020multiple} are based on multiple teacher-student networks and achieve good performance. However, $LF^{2}$ outperforms them by 8.4\% and 7.4\% in mAP accuracy respectively, which uses only a pair of teacher-student network. 
On \textbf{D-to-M}, $LF^{2}$ achieves an mAP of 1.7\% improvement compared with other fine-tuning methods. 
Based on joint learning method, IDM\cite{dai2021idm} uses domain specific batch normalization and achieves best Rank-1 accuracy. RDSBN+MDIF\cite{bai2021unsupervised} and $P^{2}LR$\cite{han2021delving} methods also construct teacher-student networks as baseline. However, these fine-tuning methods only focus on either pseudo labels refinery or domain-level information fusion. Compared with them, our $LF^{2}$ framework takes full use of global and local features to learn more comprehensive representations for clustering.

\begin{figure}[t]\centering
	\includegraphics[scale=0.14]{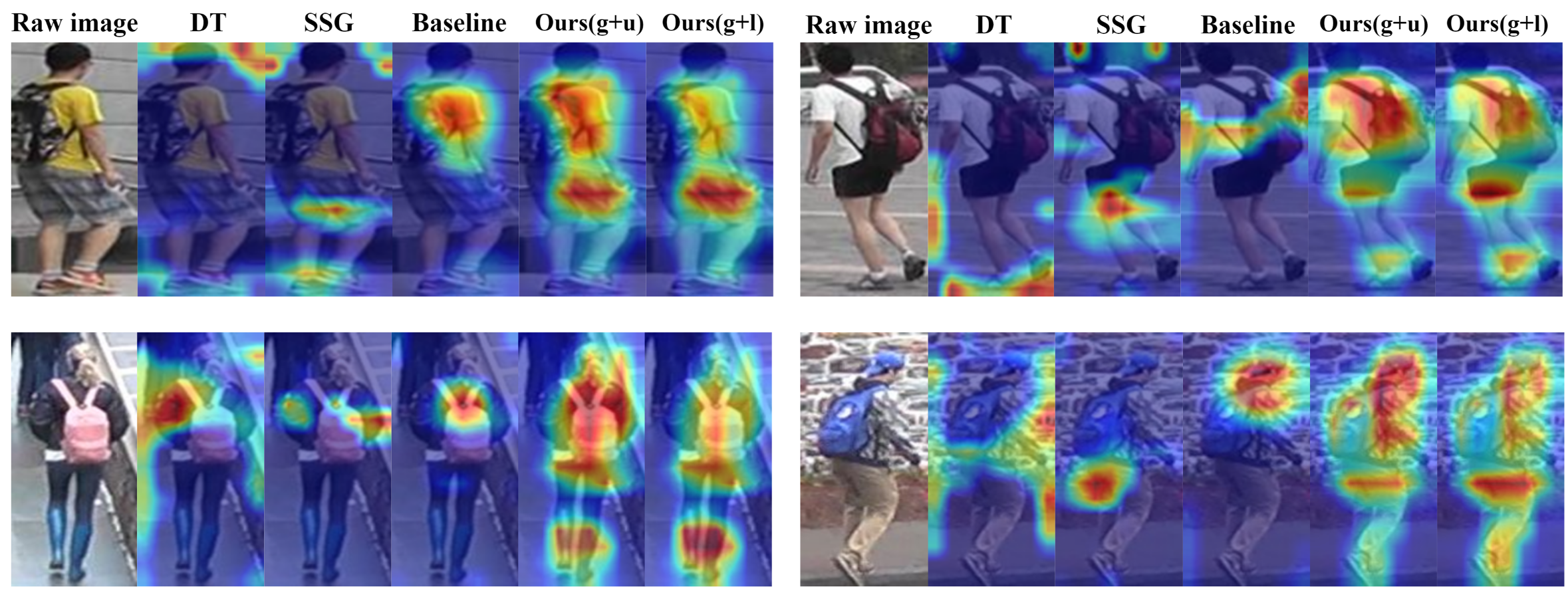}
	\caption{Examples of grad-CAMs\cite{selvaraju2017grad} of pedestrians. With our $LF^{2}$ framework, the grad-CAMs focus on both global and local information.}
	\label{FIG:4}
\end{figure}

\begin{figure}[]\centering
	\includegraphics[scale=0.16]{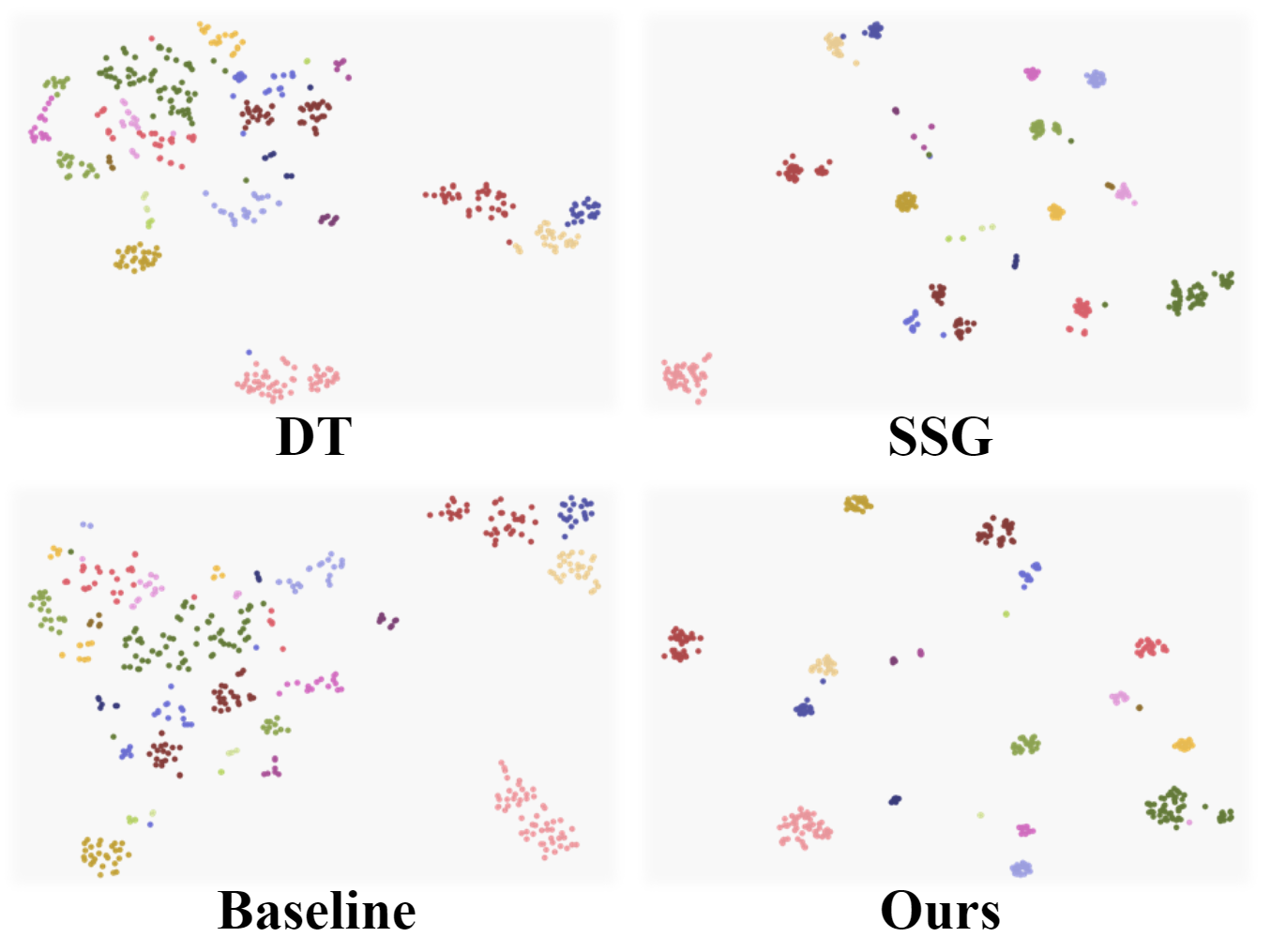}
	\caption{t-SNE\cite{van2008visualizing} visualization of 20 pedestrians on target domain. Different colors indicate different identities. Best viewed in color and zoom in.}
	\label{FIG:5}
\end{figure}

\subsection{Ablation Study}
\textbf{Comparison of Baseline Method.} 
We create a baseline, termed as ``\textbf{Baseline (only \bm{$L_{ReID}^{t}$})}'' in Table \ref{table_2}. It only uses the teacher network's global feature for clustering and then supervises the corresponding of the student network with the generated pseudo labels $\hat{Y}_{0}$. ``\textbf{Direct transfer}'' indicates that directly using the source-domain pre-trained model to adapt the target domain. Despite the baseline with the teacher-student mode brings significant improvement, $LF^{2}$ outperforms the baseline by a large margin, which proves that the local feature is able to improve the adaptation capability. 

\textbf{Effectiveness of \bm{$FM$}.} 
Experiments of removing $FM$ but keeping $L_{tri}^{P_{j}}$ are conducted, which are denoted as ``\textbf{\bm{$LF^{2}$} w/o \bm{$FM$}}'' in Table \ref{table_2}. ``w/o'' is short for without. 
Specifically, we horizontally split the global feature map of the teacher network and use the split features and the global features for clustering, as shown by the dotted line in Fig.\ref{FIG:2}.
Evident improvements of up to 9.5\% and 7.2\% in mAP accuracy of ``$LF^{2}$ w/o $FM$'' compared to the baseline method on D-to-M and M-to-D. With a learnable $FM$, $LF^{2}$ achieves 4.7\% and 5.0\% improvement in mAP accuracy on D-to-M and M-to-D when $M_{t,j}$ is set to 700 and 900 respectively.

\textbf{Evaluation on the number \bm{$M_{t,j}$} of pseudo identities.} 
We utilize K-means algorithm and the number $M_{t,j}$ is set as 500, 700 and 900, respectively. These values are either smaller or greater than the true number of identities. Note that $M_{t,j}$ of global and fusion features are set as the same values, where $j \in \{0,1,2\}$. As shown in Table \ref{table_2}, $LF^{2}$ performs best on D-to-M and M-to-D when $M_{t,j}$ are 700 and 900, respectively.

\textbf{Visualization of feature maps.} 
To verify the reliability of fusion features obtained by $LF^{2}$, we visualize the fusion feature maps by grad-CAMs\cite{selvaraju2017grad}, as shown in Fig.\ref{FIG:4}. ``\textbf{Ours(g+u)}'' and ``\textbf{Ours(g+l)}'' denote the visualization results of fusing the upper and lower feature maps of the student network with the global feature map of the teacher network, respectively. For a comparison, we also visualize the feature maps of the last convolutional layer of the ResNet50 backbone for other methods, as shown in Fig.\ref{FIG:4}.
 ``\textbf{DT}'' denotes Direct Transfer method. 
Fig.\ref{FIG:4} indicates that $LF^{2}$ pays great attention to both global and local information, \textit{e.g.}, the person in the upper left corner (yellow clothes, black schoolbag, plaid shorts) and the person in the lower right corner (blue hat and schoolbag, spotted sleeve, beige trousers). This shows the validity of using $LF^{2}$ for learning more comprehensive representations. 

\begin{figure}[t]\centering
	\includegraphics[scale=0.11]{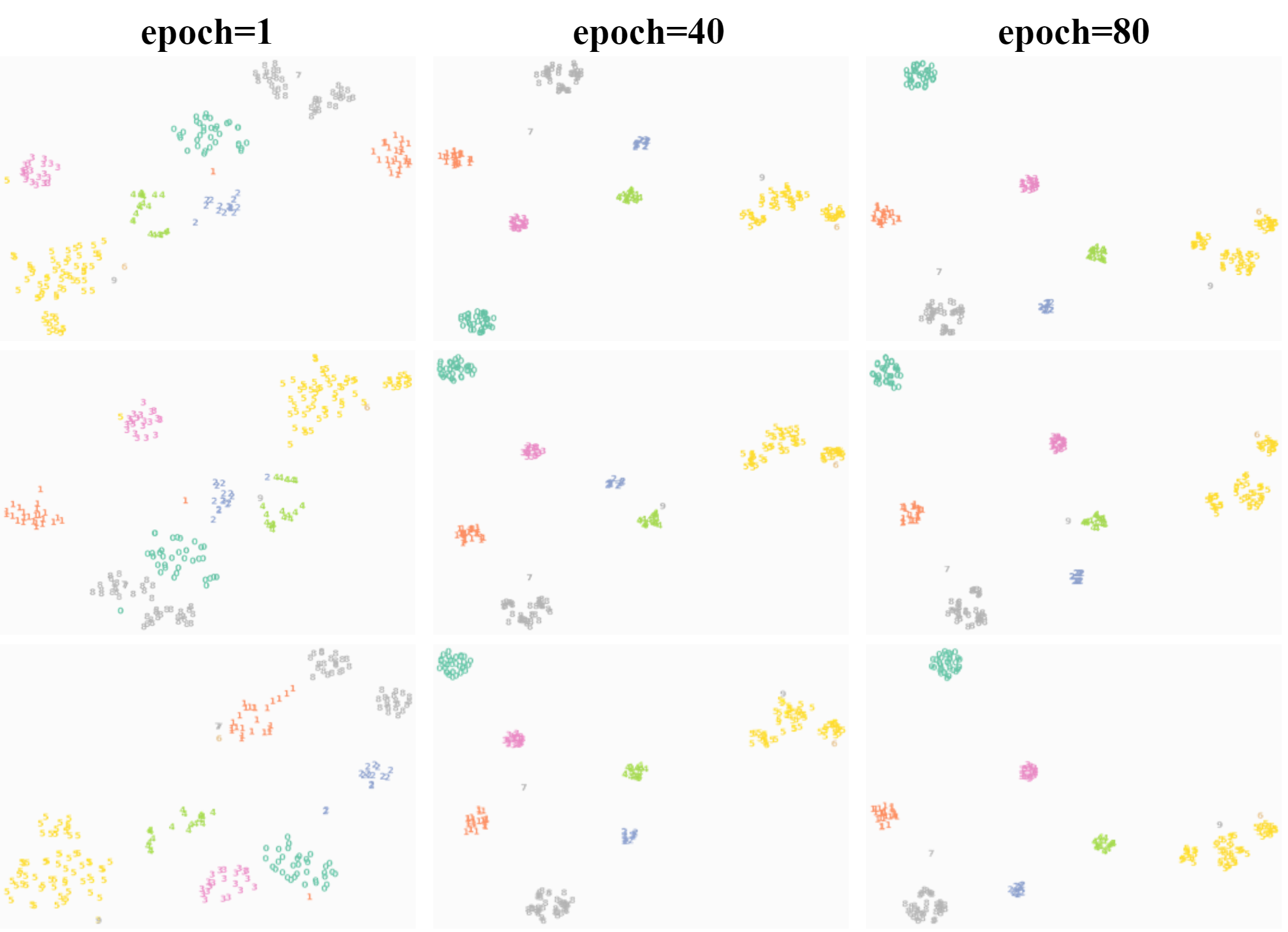}
	\caption{Visualization of 10 pedestrians during target-domain fine-tuning with our $LF^{2}$ framework. Each row denotes the results of different features. From top to bottom is $\varphi^{P_{0}}$,$\varphi^{P_{1}}$ and $\varphi^{P_{2}}$. The consistent results of these three features avoid obscure learning. }
	\label{FIG:6}
\end{figure}

\textbf{Visualization of clustering features.} 
We use t-SNE\cite{van2008visualizing} to reduce the global features of four different methods into a 2-D space, as shown in Fig.\ref{FIG:5}. 
SSG tends to pull the feature of each sample more separate due to the obscure learning even if some samples belong to the same identity. Compared with \textbf{DT} and \textbf{Baseline}, $LF^{2}$ pushes the features of the same identity more compact. This suggests that $LF^{2}$ can learn more comprehensive representations for clustering. Fig.\ref{FIG:6} illustrates the clustering process of the global and fusion features during the target-domain fine-tuning. The clusters of global and fusion features gradually become consistent, which means that $LF^{2}$ avoids obscure learning during training. 

\section{Conclusion}
In this paper, we propose a Learning Feature Fusion ($LF^{2}$) framework that adaptively learns to fuse global and local features to obtain more comprehensive fusion feature representations. A learnable Fusion Module ($FM$) is also proposed to avoid obscure learning of multiple pseudo labels. 
Visualizations of feature maps and clustering features have demonstrated the capability of adaptively learning fusion and the effectiveness of $FM$. 
We hope that our work can provide a new insight for UDA person ReID. 

\bibliographystyle{IEEEtran}

\clearpage
\bibliography{main-refs}

\clearpage

\renewcommand\thetable{\Alph{table}}
\end{document}